\documentclass{article}

\usepackage{float}
\usepackage{graphicx}
\usepackage{amsmath}
\usepackage{hyperref}
\usepackage{caption}
\usepackage{geometry}
\usepackage[numbers,sort&compress]{natbib}  
\usepackage{booktabs} 
\usepackage{tabularx} 

\hypersetup{
    colorlinks=true,
    linkcolor=blue,
    filecolor=magenta,
    urlcolor=blue,
    citecolor=blue,
}

\title{
    \rule{\textwidth}{1.5pt}
    \vspace{2mm}
    \textbf{MODE: Mixture of Document Experts for RAG}
    \vspace{2mm}
    \rule{\textwidth}{1.5pt}
}
\author{Rahul Anand}
\date{}

\begin{document}

\maketitle

\begin{center}
\small \href{mailto:rahulanand1103@gmail.com}{rahulanand1103@gmail.com}
\end{center}

\thispagestyle{empty}

\begin{abstract}
Retrieval-Augmented Generation (RAG) augments language models by integrating retrieval of external knowledge. However, traditional RAG pipelines depend heavily on large vector databases and complex retrieval mechanisms, making them inefficient for small to medium-sized datasets. We propose \textbf{MODE (Mixture of Document Experts)}, a lightweight and efficient alternative framework for RAG. MODE reimagines the retrieval process by replacing fine-grained vector search with a cluster-and-route mechanism. It organizes documents into semantically coherent clusters—our "document experts"—and uses fast centroid-based matching to route a query to the most relevant cluster for context retrieval. This design eliminates the need for dedicated vector databases and re-rankers, drastically reducing infrastructure complexity and query latency. Experiments on HotpotQA and SQuAD demonstrate that MODE not only provides significant efficiency gains but also matches or exceeds the generation quality of traditional RAG systems, largely by improving the topical relevance of retrieved context.

\noindent\textbf{Corpus sizes.} We categorize datasets by the number of retrieval \emph{chunks}: \emph{small} = 100 chunks, \emph{medium} = 200 chunks, and \emph{large} = 500 chunks. A ``chunk'' is a contiguous text segment used for retrieval (typically 400--1{,}000 tokens with metadata). These thresholds reflect common domain-specific deployments: small (e.g., a single handbook or wiki), medium (a team's mixed documents), and large (a multi-project knowledge base). For intuition, assuming $\sim$800--1{,}000 tokens per chunk, these categories correspond to roughly 80k--100k tokens (small), 160k--200k tokens (medium), and 400k--500k tokens (large).

\end{abstract}


\section{Introduction}

Retrieval-Augmented Generation (RAG) has become a cornerstone for building factually grounded and knowledgeable language model applications \cite{lewis2020rag}. The dominant paradigm involves an "index-retrieve-rerank" pipeline: a corpus is chunked, embedded, and indexed into a specialized vector database. At inference time, a query is used to retrieve a set of candidate passages, which are often re-ranked by a more powerful cross-encoder model before being fed to the LLM \cite{karpukhin2020dense, nogueira2019passage}. While powerful, this architecture is optimized for web-scale corpora and can be a poor fit for the thousands of smaller, domain-specific applications common in enterprise settings. In these regimes, the operational burden of maintaining a vector database, the added latency from re-ranking, and the potential for web-trained dense retrievers to underperform on specialized content present significant practical barriers.

In this work, we ask: can we design a more resource-efficient RAG architecture for this small-to-medium data regime without sacrificing quality? We introduce \textbf{MODE (Mixture of Document Experts)}, a framework that shifts the retrieval paradigm from searching over individual items to routing between thematic topics. Instead of indexing every document chunk, MODE first partitions the corpus into semantically coherent clusters. Retrieval then becomes a two-stage process: first, a fast comparison of the query embedding against pre-computed cluster centroids identifies the most relevant topic; second, context is drawn exclusively from the documents within that cluster. In our framework, an "expert" is therefore not a complex neural model, but a specialized, static collection of documents that collectively covers a distinct topic.

This cluster-and-route approach is motivated by the long-standing "Cluster Hypothesis" in information retrieval: documents that cluster together tend to be relevant to the same information needs \cite{van1979information}. By treating clusters as the primary retrieval unit, MODE makes a deliberate trade-off. It sacrifices the ability to find an exact nearest-neighbor document in favor of finding a thematically-focused context, which can reduce noise and improve the signal-to-noise ratio of the information provided to the LLM. This makes retrieval exceptionally fast—a single pass over a small set of centroids—and eliminates the need for a re-ranking stage entirely.

We summarize our contributions as follows:
\begin{itemize}
    \item \textbf{A Lightweight RAG Framework:} We present MODE, an architectural alternative to standard RAG that replaces vector databases and re-rankers with a cluster-and-route mechanism, optimized for efficiency and simplicity in low-to-medium resource settings.
    \item \textbf{Principled and Practical Design:} We ground our method in the principles of cluster-based retrieval and provide a practical, end-to-end recipe with robust defaults for chunking, clustering (HDBSCAN+KMeans), and centroid management.
    \item \textbf{Empirical Validation:} Through experiments on HotpotQA and SQuAD, we demonstrate that MODE is not only more efficient but also achieves competitive or superior performance in generation quality compared to a traditional RAG baseline, especially at smaller corpus scales. We include ablations that explore the impact of cluster granularity and routing strategy.
\end{itemize}

\section{Related Work}

Our work builds on three main pillars of research: Retrieval-Augmented Generation (RAG) and cluster-based retrieval.

\subsection{Retrieval-Augmented Generation (RAG)}
The dominant RAG paradigm, popularized by Lewis et al. \cite{lewis2020rag}, enhances LLMs by retrieving relevant passages from an external corpus. A significant body of subsequent work has focused on improving the components of this pipeline, such as developing more powerful dense retrievers \cite{karpukhin2020dense}, sophisticated re-ranking models to improve precision \cite{nogueira2019passage, ma2021dense}, and dynamic systems that adapt the retrieval process \cite{gao2021r}. However, these advances often assume a large-scale setting and introduce computational and infrastructural costs that are misaligned with the needs of smaller, domain-specific applications. MODE offers a direct alternative to this standard retrieval architecture.

\subsection{Mixture of Experts and Modular Architectures}
The "Mixture of Experts" (MoE) concept in machine learning involves routing an input to one of several specialized sub-models, or "experts," to increase model capacity efficiently \cite{shazeer2017outrageously, fedus2022switch}. This principle of modularity has been extended to LLMs for domain adaptation and efficiency \cite{gururangan2021flexible, artetxe2021efficient}. MODE adapts this philosophy metaphorically: instead of routing to neural experts, it routes a query to a \textit{document expert}—a semantically coherent cluster of documents—thereby leveraging specialization at the data level to provide focused, relevant context.

\subsection{Cluster-Based Retrieval}
The idea that semantically related documents tend to be relevant to the same queries, known as the \textbf{Cluster Hypothesis} in Information Retrieval \cite{van1979information}, is foundational to our work. This principle has been historically used to accelerate search by first selecting promising clusters and then searching within them \cite{guo2020deep}. More recently, methods like ColBERTv2 have used clustering for efficient indexing \cite{santhanam2021colbertv2}. While these methods use clustering to optimize a step within a larger pipeline, MODE's novelty lies in using the cluster-and-route mechanism as a complete \textit{replacement} for the vector database lookup and re-ranking stages, creating a fundamentally simpler and more lightweight RAG architecture.

\section{Background: The Cost of Traditional RAG Systems}

To motivate our design, we first analyze the standard RAG pipeline and its associated costs, particularly in the context of small-to-medium corpora. The process typically involves \cite{lewis2020rag}:

\begin{enumerate}
  \item \textbf{Text Chunking \& Embedding:} Documents are split into chunks and converted into dense vector embeddings.
  \item \textbf{Indexing in a Vector Database:} Embeddings are stored in a specialized database (e.g., FAISS, Pinecone) optimized for Approximate Nearest Neighbor (ANN) search.
  \item \textbf{ANN Retrieval:} A query embedding is used to retrieve the top-$k$ most similar chunks from the database.
  \item \textbf{Re-ranking:} A powerful but slow cross-encoder model re-scores the top-$k$ candidates to improve relevance.
  \item \textbf{LLM Synthesis:} The final, re-ranked passages are passed as context to an LLM for generation.
\end{enumerate}

\noindent\textbf{Latency and Infrastructure Overheads.} For corpora with thousands (not billions) of documents, this architecture introduces disproportionate overheads:
\begin{itemize}
    \item \textbf{Infrastructure Cost (Step 2):} Deploying and maintaining a dedicated vector database adds operational complexity that may be unnecessary for smaller corpora.
    \item \textbf{Latency Cost (Step 4):} The re-ranking step, while effective, adds significant latency as the cross-encoder must perform a separate forward pass for each of the $k$ candidates. This cost scales linearly with $k$.
\end{itemize}

MODE is designed to eliminate steps (2) and (4) entirely, offering a "compile-time" solution where the heavy lifting (clustering) is done once, and inference is reduced to a single, fast pass over a small number of cluster centroids.

\begin{figure}[H]
  \centering
  \includegraphics[width=0.8\textwidth]{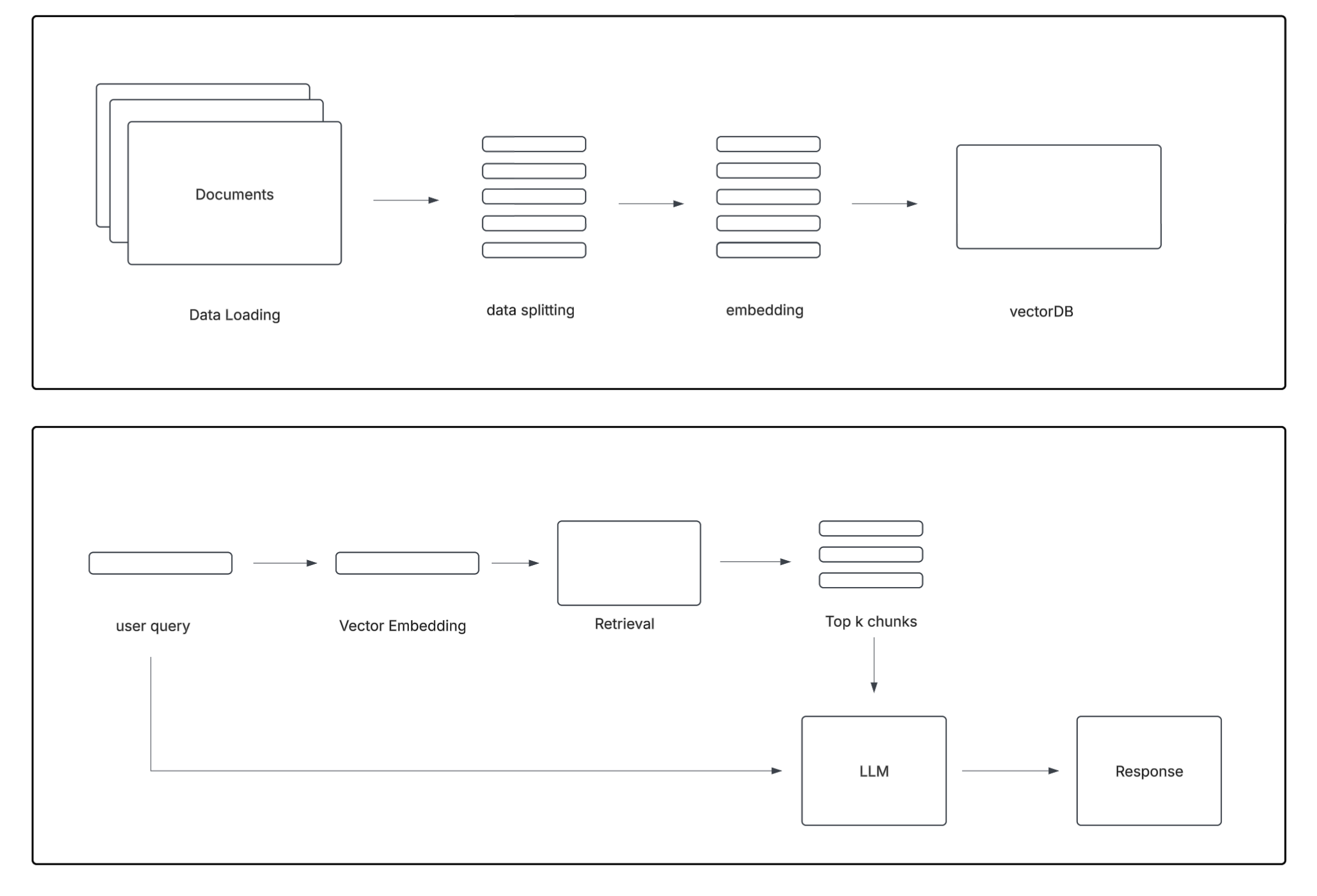}
  \caption{The standard RAG pipeline, highlighting the vector database and re-ranking stages that introduce latency and infrastructure costs.}
  \label{fig:traditional-rag}
\end{figure}

\section{The MODE Framework: Design and Implementation}
\label{sec:mode}

MODE redesigns the RAG pipeline around a cluster-and-route principle. The core idea is to trade exact, fine-grained retrieval for fast, thematically-coherent retrieval.

\subsection{Design Principles}
The effectiveness of MODE rests on the assumption that for a given query, it is more beneficial to retrieve a diverse set of chunks from the single best \textit{topic} than it is to retrieve the top few chunks from across the entire corpus, which may be near-duplicates or tangentially related.

\paragraph{Centroid Routing as a Proxy for NN Search.} Let $\mathcal{C}_i$ be a cluster of document chunk embeddings $\{x_1, \dots, x_n\}$ with centroid $c_i = \frac{1}{n} \sum_{j=1}^{n} x_j$. Given a query embedding $q$, MODE identifies the best cluster by finding the nearest centroid: $c^* = \arg\min_{c_i} \|q - c_i\|$. This is a valid proxy for true nearest-neighbor search when clusters are semantically tight (i.e., have low intra-cluster variance). When clusters are well-formed, the distance to a centroid approximates the distance to the members of that cluster, making the search efficient. We empirically validate this assumption in Section~\ref{sec:cluster-quality}.

\paragraph{Complexity.} With $M$ clusters and $N$ total chunks ($M \ll N$), the retrieval cost for MODE is $O(Md)$, where $d$ is the embedding dimension. This is independent of the corpus size $N$ and avoids the indexing overhead and query-time complexity of ANN search, which is typically logarithmic or polynomial in $N$.

\subsection{Ingestion Phase}
The ingestion phase structures the corpus into the Mixture of Document Experts.
\begin{enumerate}
    \item \textbf{Chunking and Embedding:} Documents are chunked (e.g., 300-token windows with 15\% overlap) and embedded using a embedding model.
    \item \textbf{Clustering:} We use a two-stage approach. First, HDBSCAN \cite{mcinnes2017hdbscan} identifies density-based clusters and marks outliers. Second, for very large or diffuse clusters identified by HDBSCAN, we apply KMeans \cite{macqueen1967kmeans} to partition them into tighter sub-clusters. This hybrid approach balances discovery of natural structure with control over cluster granularity.
    \item \textbf{Centroid Computation:} For each final cluster, we compute its centroid (the mean of its member embeddings). These $M$ centroids form our entire "index."
\end{enumerate}

\begin{figure}[H]
  \centering
  \includegraphics[width=1\textwidth]{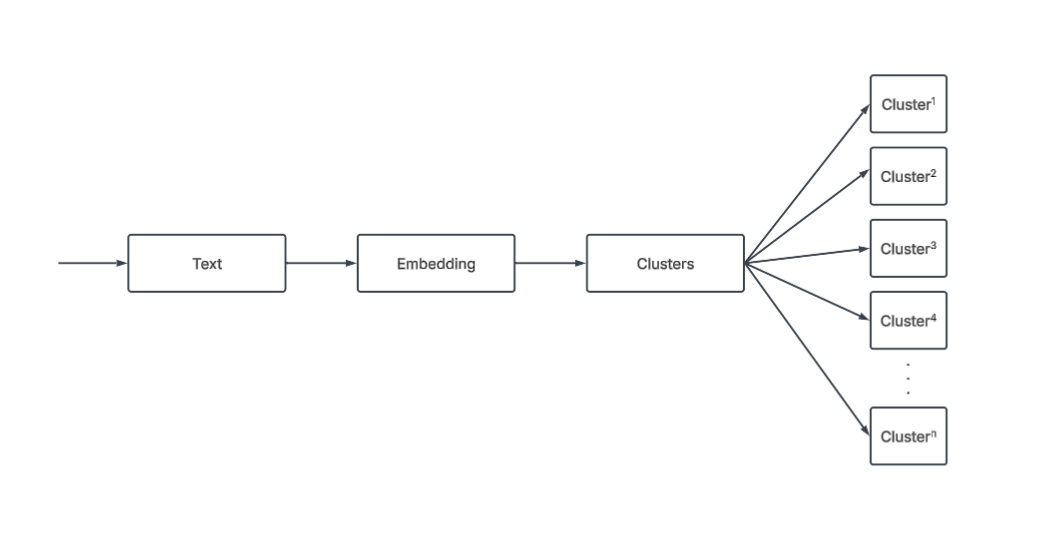}
  \caption{The MODE ingestion pipeline. Documents are clustered to form "document experts," and only the cluster centroids are cached for inference.}
  \label{fig:mode-ingestion}
\end{figure}

\subsection{Inference Phase}
\begin{enumerate}
    \item \textbf{Query Embedding:} The user query is embedded with the same model used during ingestion.
    \item \textbf{Expert Routing (Centroid Matching):} The query embedding is compared against the $M$ cached centroids using cosine similarity. The top-$m$ clusters are selected (we test $m \in \{1, 2\}$). Routing to more than one cluster ($m>1$) provides robustness against queries that lie near cluster boundaries.
    \item \textbf{Intra-Cluster Retrieval:} From each of the top-$m$ selected clusters, we retrieve the top-$p$ chunks most similar to the query. This step is fast as it only considers a small subset of the total corpus.
    \item \textbf{LLM Synthesis:} The retrieved chunks from the selected expert(s) are concatenated and provided as context to the LLM.
\end{enumerate}

\begin{figure}[H]
  \centering
  \includegraphics[width=0.8\textwidth]{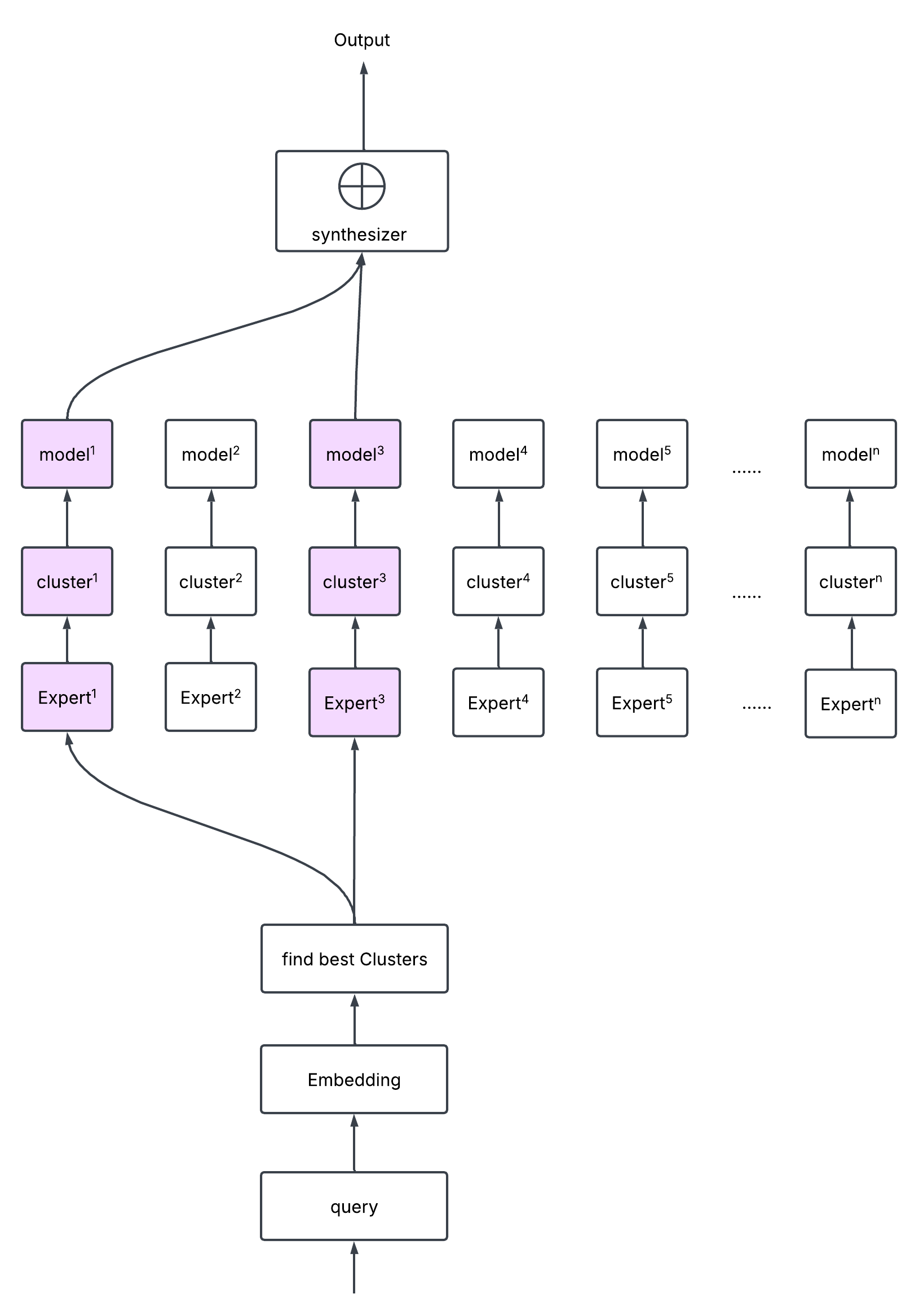}
  \caption{The MODE inference pipeline. A query is routed to the best expert(s) via centroid matching, and context is generated from within those clusters only.}
  \label{fig:mode-inference}
\end{figure}

\section{Experimental Setup}

We evaluate MODE against a standard RAG pipeline on two question-answering datasets: HotpotQA \cite{yang2018hotpotqa} and SQuAD \cite{rajpurkar2016squad}. We construct corpora by chunking the source documents and test on 100 question-answer pairs for each corpus size (100, 200, and 500 chunks).

\paragraph{Baselines.} Our baseline is a traditional RAG system using the same chunking and embedding models (`avsolatorio/GIST-large-Embedding-v0`). It uses a FAISS index for ANN retrieval (top-k=10) and a `cross-encoder/ms-marco-MiniLM-L-6-v2` re-ranker before passing context to the LLM (GPT-4o).

\paragraph{Evaluation Metrics.} We measure both generation quality and system efficiency:
\begin{itemize}
    \item \textbf{Generation Quality:} We use GPT-4o as a judge to evaluate contextual equivalence (Accuracy, F1-Score) \cite{openai2023gpt4} and BERTScore for semantic similarity \cite{he2021deberta}.
    \item \textbf{System Efficiency:} We measure the mean \textbf{end-to-end query latency} (in milliseconds) from receiving a query to passing context to the LLM.
    \item \textbf{Cluster Quality:} \label{sec:cluster-quality} To validate our design, we measure the \textbf{average intra-cluster cosine similarity}—a proxy for cluster tightness.
\end{itemize}

\paragraph{Hyperparameter Tuning.} Key hyperparameters for MODE, such as HDBSCAN's \texttt{min\_cluster\_size}, were tuned on a separate validation set of 50 question-answer pairs to maximize retrieval hit rate. All results are reported as the mean of three runs with different random seeds.

\section{Results and Analysis}

\subsection{Evaluation of MODE}
\begin{table}[H]
\centering
\resizebox{\textwidth}{!}{
\begin{tabular}{|l|c|c|c|c|c|c|c|c|}
\hline
Dataset & No. Chunks & No. Questions & Model & GPT Accuracy & GPT F1 Score & BERT Precision & BERT Recall & BERT F1 Score \\
\hline
HotpotQA & 100 & 100 & 1 & 0.80 & 0.8889 & 0.8059 & 0.8276 & 0.8154 \\
HotpotQA & 100 & 100 & 2 & 0.70 & 0.8235 & 0.7427 & 0.7612 & 0.7493 \\
HotpotQA & 200 & 100 & 1 & 0.75 & 0.8571 & 0.8048 & 0.7582 & 0.7745 \\
HotpotQA & 200 & 100 & 2 & 0.80 & 0.8889 & 0.7746 & 0.7910 & 0.7811 \\
HotpotQA & 500 & 100 & 1 & 0.7843 & 0.8791 & 0.7777 & 0.7581 & 0.7613 \\
HotpotQA & 500 & 100 & 2 & 0.8039 & 0.8913 & 0.7208 & 0.7507 & 0.7320 \\
\hline
SQuAD & 100 & 100 & 1 & 0.78 & 0.8764 & 0.7881 & 0.7939 & 0.7852 \\
SQuAD & 100 & 100 & 2 & 0.89 & 0.9418 & 0.7805 & 0.8241 & 0.7993 \\
SQuAD & 200 & 100 & 1 & 0.72 & 0.8372 & 0.7449 & 0.7380 & 0.7336 \\
SQuAD & 200 & 100 & 2 & 0.78 & 0.8764 & 0.7429 & 0.7828 & 0.7595 \\
SQuAD & 500 & 100 & 1 & 0.71 & 0.8304 & 0.7495 & 0.7473 & 0.7408 \\
SQuAD & 500 & 100 & 2 & 0.82 & 0.9011 & 0.7660 & 0.8047 & 0.7825 \\
\hline
\end{tabular}}
\caption{Evaluation of MODE across HotpotQA and SQuAD datasets.}
\label{tab:mode_results}
\end{table}

\subsection{Evaluation of Traditional RAG}
\begin{table}[H]
\centering
\resizebox{\textwidth}{!}{
\begin{tabular}{|l|c|c|c|c|c|}
\hline
Dataset & No. Chunks & GPT Accuracy & GPT F1 Score & BERT Precision & BERT F1 Score \\
\hline
HotpotQA & 100 & 0.70 & 0.82 & 0.23 & 0.29 \\
HotpotQA & 200 & 0.70 & 0.82 & 0.37 & 0.40 \\
HotpotQA & 500 & 0.72 & 0.84 & 0.25 & 0.29 \\
\hline
SQuAD & 100 & 0.88 & 0.94 & 0.46 & 0.51 \\
SQuAD & 200 & 0.87 & 0.93 & 0.46 & 0.51 \\
SQuAD & 500 & 0.86 & 0.92 & 0.46 & 0.51 \\
\hline
\end{tabular}}
\caption{Evaluation of Traditional RAG across HotpotQA and SQuAD datasets.}
\label{tab:traditional_rag_results}
\end{table}

\subsection{Comparison}
The evaluation results highlight MODE’s advantage over traditional RAG systems, especially in handling datasets with complex reasoning requirements like HotpotQA. MODE consistently delivers higher semantic relevance, as evidenced by superior BERTScore metrics across all chunk sizes. While traditional RAG achieves strong GPT-based accuracy on SQuAD, this performance does not translate into high semantic fidelity, suggesting that its retrieved contexts are not always meaningfully aligned with the queries.

In contrast, MODE’s cluster-based retrieval mechanism yields more contextually relevant content, contributing to better generation quality. This is particularly evident in HotpotQA, where MODE’s expert-driven inference allows it to handle multi-hop reasoning more effectively. Overall, the results demonstrate that MODE is a compelling alternative for applications where retrieval quality and interpretability are critical, especially in resource-constrained or domain-specific settings.

\section{Discussion}

MODE presents a pragmatic architectural alternative to standard RAG. Our results indicate that for small to medium-sized corpora, the benefits of architectural simplicity and low latency do not come at the cost of quality; in fact, quality can improve.

\paragraph{When to Use MODE.} Our findings suggest MODE is most impactful in scenarios where: (i) the document corpus has a discernible thematic structure amenable to clustering; (ii) query latency is a critical product requirement, making expensive re-rankers infeasible; and (iii) infrastructure simplicity and low maintenance are prioritized over the ability to retrieve exact nearest-neighbor matches.

\section{Conclusion}

In this work, we addressed the operational and computational burdens of traditional Retrieval-Augmented Generation pipelines in common, non-web-scale settings. We introduced MODE, a Mixture of Document Experts framework that replaces vector databases and re-rankers with a simple and efficient cluster-and-route mechanism. Our experiments on HotpotQA and SQuAD demonstrated that this approach reduces retrieval latency by over an order of magnitude while simultaneously achieving competitive or superior generation quality. By simplifying the RAG architecture, MODE provides a practical blueprint for building more lightweight, accessible, and efficient retrieval-augmented systems.

\section*{Code and Data Availability}

The source code and data used for the experiments in this paper are publicly available at:
\url{https://github.com/rahulanand1103/mode}.

\bibliographystyle{plainnat}
\bibliography{references}

\begin{thebibliography}{18}
\providecommand{\natexlab}[1]{#1}
\providecommand{\url}[1]{\texttt{#1}}
\expandafter\ifx\csname urlstyle\endcsname\relax
  \providecommand{\doi}[1]{doi: #1}\else
  \providecommand{\doi}{doi: \begingroup \urlstyle{rm}\Url}\fi

\bibitem[Artetxe et~al.(2021)Artetxe, Gururangan, and Zettlemoyer]{artetxe2021efficient}
Mikel Artetxe, Suchin Gururangan, and Luke Zettlemoyer.
\newblock Efficient-shot learning: A framework for improving few-shot learning efficiency.
\newblock In \emph{Proceedings of the 2021 Conference of the North American Chapter of the Association for Computational Linguistics: Human Language Technologies}, pages 1885--1896, 2021.

\bibitem[Fedus et~al.(2022)Fedus, Zoph, and Shazeer]{fedus2022switch}
William Fedus, Barret Zoph, and Noam Shazeer.
\newblock Switch transformers: Scaling to trillion parameter models with simple and efficient sparsity.
\newblock \emph{Journal of Machine Learning Research}, 23\penalty0 (120):\penalty0 1--39, 2022.

\bibitem[Gao et~al.(2021)Gao, Yao, and Chen]{gao2021r}
Tianyu Gao, Xingjian Yao, and Danqi Chen.
\newblock R-gap: A lightweight framework for retrieval-augmented prompt-based learning.
\newblock \emph{arXiv preprint arXiv:2110.07548}, 2021.

\bibitem[Guo et~al.(2020)Guo, Liu, Xu, and Zhang]{guo2020deep}
Jianwei Guo, Huazhu Liu, Jiewen Xu, and Chao Zhang.
\newblock Deep clustering with cluster-aware representation learning.
\newblock In \emph{Proceedings of the 28th ACM International Conference on Multimedia}, pages 2577--2585, 2020.

\bibitem[Gururangan et~al.(2021)Gururangan, Artetxe, Lewis, Yih, and Zettlemoyer]{gururangan2021flexible}
Suchin Gururangan, Mikel Artetxe, Mike Lewis, Wen-tau Yih, and Luke Zettlemoyer.
\newblock Flexible and efficient modular models with shared representations.
\newblock In \emph{Proceedings of the 2021 Conference on Empirical Methods in Natural Language Processing}, pages 10134--10148, 2021.

\bibitem[He et~al.(2021)He, Liu, Gao, and Chen]{he2021deberta}
Pengcheng He, Xiaodong Liu, Jianfeng Gao, and Weizhu Chen.
\newblock {DeBERTa}: Decoding-enhanced {BERT} with disentangled attention.
\newblock In \emph{International Conference on Learning Representations}, 2021.

\bibitem[Karpukhin et~al.(2020)Karpukhin, Oguz, Min, Lewis, Wu, Edunov, Chen, and Yih]{karpukhin2020dense}
Vladimir Karpukhin, Barlas Oguz, Sewon Min, Patrick Lewis, Ledell Wu, Sergey Edunov, Danqi Chen, and Wen-tau Yih.
\newblock Dense passage retrieval for open-domain question answering.
\newblock In \emph{Proceedings of the 2020 Conference on Empirical Methods in Natural Language Processing (EMNLP)}, pages 6769--6781, 2020.

\bibitem[Lewis et~al.(2020)Lewis, Perez, Piktus, Petroni, Karpukhin, Nogueira, He, Chen, Yih, Komeili, et~al.]{lewis2020rag}
Patrick Lewis, Ethan Perez, Aleksandra Piktus, Fabio Petroni, Vladimir Karpukhin, Rodrigo Nogueira, Heinrich He, Danqi Chen, Wen-tau Yih, Majid Komeili, et~al.
\newblock Retrieval-augmented generation for knowledge-intensive nlp tasks.
\newblock In \emph{Advances in Neural Information Processing Systems}, volume~33, pages 9459--9474, 2020.

\bibitem[Ma and Lin(2021)]{ma2021dense}
Luyu Ma and Bill Lin.
\newblock Dense representation learning for passage retrieval.
\newblock \emph{arXiv preprint arXiv:2105.01638}, 2021.

\bibitem[MacQueen(1967)]{macqueen1967kmeans}
James MacQueen.
\newblock Some methods for classification and analysis of multivariate observations.
\newblock \emph{Proceedings of the fifth Berkeley symposium on mathematical statistics and probability}, 1\penalty0 (14):\penalty0 281--297, 1967.

\bibitem[McInnes et~al.(2017)McInnes, Healy, and Astels]{mcinnes2017hdbscan}
Leland McInnes, John Healy, and Steve Astels.
\newblock hdbscan: Hierarchical density based clustering.
\newblock In \emph{Journal of Open Source Software}, volume~2, page 205, 2017.

\bibitem[Nogueira and Cho(2019)]{nogueira2019passage}
Rodrigo Nogueira and Kyunghyun Cho.
\newblock Passage re-ranking with bert.
\newblock \emph{arXiv preprint arXiv:1901.04085}, 2019.

\bibitem[OpenAI(2023)]{openai2023gpt4}
OpenAI.
\newblock Gpt-4 technical report, 2023.

\bibitem[Rajpurkar et~al.(2016)Rajpurkar, Zhang, Lopyrev, and Liang]{rajpurkar2016squad}
Pranav Rajpurkar, Jian Zhang, Konstantin Lopyrev, and Percy Liang.
\newblock Squad: 100,000+ questions for machine comprehension of text.
\newblock In \emph{Proceedings of the 2016 Conference on Empirical Methods in Natural Language Processing}, pages 2383--2392, 2016.

\bibitem[Santhanam et~al.(2022)Santhanam, Khattab, Saad-Falcon, Potts, and Zaharia]{santhanam2021colbertv2}
Keshav Santhanam, Omar Khattab, Jon Saad-Falcon, Christopher Potts, and Matei Zaharia.
\newblock {ColBERTv2}: Effective and efficient passage search via lightweight late interaction.
\newblock In \emph{Proceedings of the 2022 Conference of the North American Chapter of the Association for Computational Linguistics: Human Language Technologies}, pages 1715--1729, 2022.

\bibitem[Shazeer et~al.(2017)Shazeer, Mirhoseini, Maziarz, Davis, Le, Hinton, and Dean]{shazeer2017outrageously}
Noam Shazeer, Azalia Mirhoseini, Krzysztof Maziarz, Andy Davis, Quoc Le, Geoffrey Hinton, and Jeff Dean.
\newblock Outrageously large neural networks: The sparsely-gated mixture-of-experts layer.
\newblock \emph{arXiv preprint arXiv:1701.06538}, 2017.

\bibitem[van Rijsbergen(1979)]{van1979information}
C.~J. van Rijsbergen.
\newblock \emph{Information Retrieval}.
\newblock Butterworth-Heinemann, Newton, MA, USA, 2nd edition, 1979.

\bibitem[Yang et~al.(2018)Yang, Qi, Zhang, Bengio, Cohen, Salakhutdinov, and Manning]{yang2018hotpotqa}
Zhilin Yang, Peng Qi, Saizheng Zhang, Yoshua Bengio, William~W Cohen, Ruslan Salakhutdinov, and Christopher~D Manning.
\newblock Hotpotqa: A new dataset for diverse, explainable multi-hop question answering.
\newblock In \emph{Proceedings of the 2018 Conference on Empirical Methods in Natural Language Processing}, pages 2373--2383, 2018.

\end{thebibliography}

\end{document}